\pdfoutput=1

\documentclass[11pt]{article}

\usepackage{EMNLP2023}

\usepackage{times}
\usepackage{latexsym}
\usepackage{graphicx}
\usepackage{svg}
\svgpath{{./figs/}}
\usepackage{booktabs}
\usepackage{threeparttable}
\usepackage{caption}
\usepackage{enumitem}
\usepackage{longtable}
\usepackage{pdflscape}
\usepackage{geometry}
\usepackage{hyperref}
\usepackage{lscape}
\usepackage{stfloats}
\usepackage{afterpage}

\usepackage[T1]{fontenc}

\usepackage[utf8]{inputenc}

\usepackage{microtype}

\usepackage{inconsolata}

\usepackage{todonotes} 
\newcommand{\ignore}[1]{\textcolor{gray}{}}

\title{An Integrative Survey on Mental Health Conversational Agents to Bridge Computer Science and Medical Perspectives}

\author{
Young-Min Cho$^1$ \quad Sunny Rai$^1$ \quad Lyle Ungar$^1$ \quad \\ \textbf{João Sedoc}$^2$ \quad \textbf{Sharath Chandra Guntuku}$^1$ \\
$^1$University of Pennsylvania \quad $^2$New York University \\
\texttt{\{jch0,sunnyrai,ungar,sharathg\}@seas.upenn.edu, jsedoc@stern.nyu.edu}
}

\begin{document}
\maketitle
\begin{abstract}
Mental health conversational agents (a.k.a. chatbots) are widely studied for their potential to offer accessible support to those experiencing mental health challenges. Previous surveys on the topic primarily consider papers published in either computer science or medicine, leading to a divide in understanding and hindering the sharing of beneficial knowledge between both domains. To bridge this gap, we conduct a comprehensive literature review using the PRISMA framework, reviewing 534 papers published in both computer science and medicine. Our systematic review reveals 136 key papers on building mental health-related conversational agents with diverse characteristics of modeling and experimental design techniques. We find that computer science papers focus on LLM techniques and evaluating response quality using automated metrics with little attention to the application while medical papers use rule-based conversational agents and outcome metrics to measure the health outcomes of participants. Based on our findings on transparency, ethics, and cultural heterogeneity in this review, we provide a few recommendations to help bridge the disciplinary divide and enable the cross-disciplinary development of mental health conversational agents. 
\end{abstract}

\section{Introduction}

The proliferation of conversational agents (CAs), also known as chatbots or dialog systems, has been spurred by advancements in Natural Language Processing (NLP) technologies. Their application spans diverse sectors, from education \cite{okonkwo2021chatbots, durall2020co} to e-commerce \cite{shenoy-etal-2021-asr}, demonstrating their increasing ubiquity and potency.

The utility of CAs within the mental health domain has been gaining recognition. Over 30\% of the world's population suffers from one or more mental health conditions; about 75\% individuals in low and middle-income countries and about 50\% individuals in high-income countries do not receive care and treatment~\cite{kohn2004treatment,arias2022quantifying}. The sensitive (and often stigmatized) nature of mental health discussions further exacerbates this problem, as many individuals find it difficult to disclose their struggles openly~\cite{corrigan2003stigma}.

Conversational agents like Woebot \cite{fitzpatrick2017delivering} and Wysa \cite{inkster2018empathy} were some of the first mobile applications to address this issue. They provide an accessible and considerably less intimidating platform for mental health support, thereby assisting a substantial number of individuals. Their effectiveness highlights the potential of mental health-focused CAs as one of the viable solutions to ease the mental health disclosure and treatment gap. 

Despite the successful implementation of certain CAs in mental health, a significant disconnect persists between research in computer science (CS) and medicine. This disconnect is particularly evident when we consider the limited adoption of advanced NLP (e.g. large language models) models in the research published in medicine. While CS researchers have made substantial strides in NLP, there is a lack of focus on the human evaluation and direct impacts these developments have on patients. Furthermore, we observe that mental health CAs are drawing significant attention in medicine, yet remain underrepresented in health-applications-focused research in NLP. This imbalance calls for a more integrated approach in future studies to optimize the potential of these evolving technologies for mental health applications.

In this paper, we present a comprehensive analysis of academic research related to mental health conversational agents, conducted within the domains of CS and medicine\footnote{Our data and papers are available on our GitHub: \href{https://github.com/JeffreyCh0/mental_chatbot_survey}{https://github.com/JeffreyCh0/mental\_chatbot\_survey}}. Employing the Preferred Reporting Items for Systematic Reviews and Meta-Analyses (PRISMA) framework \cite{moher2010preferred}, we systematically reviewed 136 pertinent papers to discern the trends and research directions in the domain of mental health conversational agents over the past five years. We find that there is a disparity in research focus and technology across communities, which is also shown in the differences in evaluation. Furthermore, we point out the issues that apply across domains, including transparency and language/cultural heterogeneity.

The primary objective of our study is to conduct a systematic and transparent review of mental health CA research papers across the domains of CS and medicine. This process aims not only to bridge the existing gap between these two broad disciplines but also to facilitate reciprocal learning and strengths sharing. In this paper, we aim to address the following key questions:
\begin{enumerate}
    \item What are the prevailing focus and direction of research in each of these domains?
    \item What key differences can be identified between the research approaches taken by each domain?
    \item How can we augment and improve mental health CA research methods?
\end{enumerate}

\section{Prior Survey Papers}
Mental health conversational agents are discussed in several non-CS survey papers, with an emphasis on their usability in psychiatry  \cite{vaidyam2019chatbots, montenegro2019survey,laranjo2018conversational}, and users' acceptability \cite{koulouri2022chatbots, gaffney2019conversational}. These survey papers 
focus on underpinning theory \cite{martinengo2022conversational}, 
standardized \textit{psychological outcomes} for evaluation \cite{vaidyam2019chatbots, gaffney2019conversational} in addition to \textit{accessibility} \cite{su2020analyzing}, \textit{safety} \cite{parmar2022health} and \textit{validity} \cite{pacheco2021smart, wilson2022development} of CAs. 

Contrary to surveys for medical audiences, NLP studies mostly focus on the quality of the generated response from the standpoint of text generation.
\citet{valizadeh2022ai} in their latest survey, reviewed $70$ articles and investigated task-oriented healthcare dialogue systems from a technical perspective. The discussion focuses on the system architecture and design of CAs.
The majority of healthcare CAs were found to have pipeline architecture despite the growing popularity of end-to-end architectures in the NLP domain. A similar technical review by \citet{safi2020technical} also reports a high reliance on static dialogue systems in CAs developed for medical applications. Task-oriented dialogue systems usually deploy a guided conversation style which fits well with rule-based systems. However, \citet{su2020analyzing, abd2021perceptions} pointed to the problem of robotic conversation style in mental health apps where users prefer an unconstrained conversation style and may even want to lead the conversation \cite{abd2019overview}.    
\citet{huang2022ideal} further underlines the need for self-evolving CAs to keep up with evolving habits and topics during the course of app usage. 

\begin{figure}[t]
    \centering
    \includegraphics[width=\linewidth]{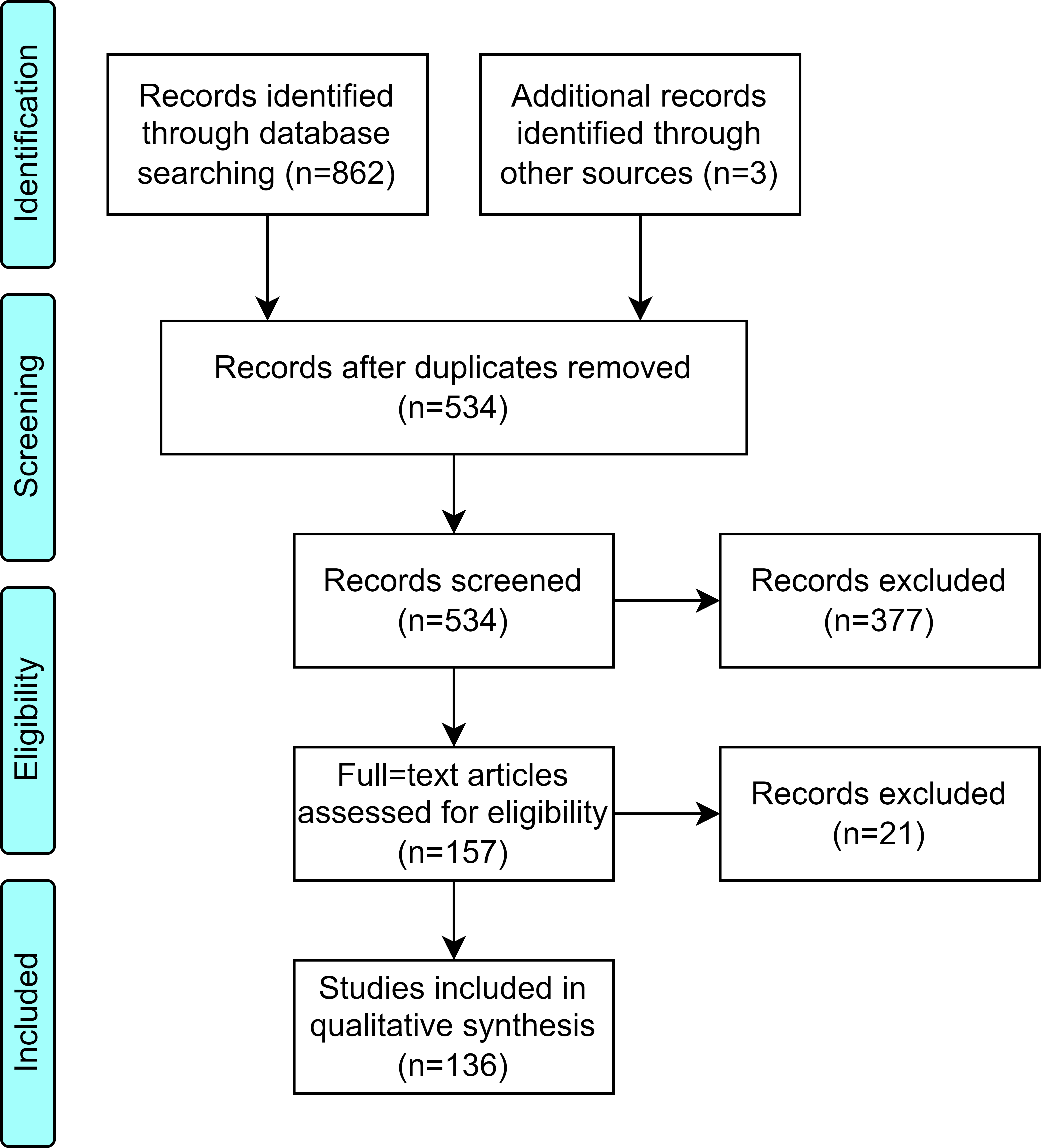}
    \caption{Pipeline of our PRISMA framework.}
    \label{fig:prisma}
\end{figure}

Surveys from the rest of CS cover HCI \cite{de2022design} and the system design of CAs \cite{dev2022comparative, narynov2021chatbots}. \citet{de2022design} analyzed 6 mental health mobile applications from an HCI perspective and suggested 24 design considerations including \textit{empathetic} conversation style, \textit{probing}, and \textit{session duration} for effective dialogue. \citet{damij2022role} proposed three key dimensions namely \textit{people} (citizen centric goals ), \textit{process} (regulations and governance) and \textit{AI technology} to consider when designing public care CAs.  

These survey papers independently provide an in-depth understanding of advancements and challenges in the CS and medical domains. However, there is a lack of studies that can provide a joint appraisal of developments to enable cross-learning across these domains. With this goal, we consider research papers from medicine (PubMed), NLP (the ACL Anthology), and the rest of CS (ACM, AAAI, IEEE) to examine the disparities in goals, methods, and evaluations of research related to mental health conversational agents.

\section{Methods}

\subsection{Paper Databases}

We source papers from eminent databases in the fields of NLP, the rest of CS, and medicine, as these are integral knowledge areas in the study of mental health CA. These databases include the ACL Anthology (referred to as ACL throughout this paper)\footnote{\url{https://aclanthology.org/}}, AAAI\footnote{\url{https://aaai.org/aaai-publications/}}, IEEE\footnote{\url{https://ieeexplore.ieee.org/}}, ACM\footnote{\url{https://dl.acm.org/}}, and PubMed\footnote{\url{https://pubmed.ncbi.nlm.nih.gov/}}. ACL is recognized as a leading repository that highlights pioneering research in NLP. AAAI features cutting-edge studies in AI. IEEE, a leading community, embodies the forefront of engineering and technology research. ACM represents the latest trends in Human Computer Interaction (HCI) along with several other domains of CS. PubMed, the largest search engine for science and biomedical topics including psychology, psychiatry, and informatics among others provides extensive coverage of the medical spectrum. 

Drawing on insights from prior literature reviews ~\cite{valizadeh2022ai, montenegro2019survey,laranjo2018conversational} and discussion with experts from both the CS and medical domains, we opt for a combination of specific keywords. These search terms represent both our areas of focus: conversational agents (``conversational agent'', ``chatbot'') and mental health (``mental health'', ``depression''). Furthermore, we limit our search criteria to the paper between 2017 to 2022 to cover the most recent articles. We also apply the ``research article'' filter on ACM search, and ``Free Full Text or Full Text'' for PubMed search. Moreover, we manually add 3 papers recommended by the domain experts \cite{fitzpatrick2017delivering, laranjo2018conversational, montenegro2019survey}. This results in 534 papers. 

\begin{table}[t]
\centering
\resizebox{\linewidth}{!}{%
\Large
\begin{tabular}{lccccc}
\toprule
\textbf{\begin{tabular}[c]{@{}l@{}}Screening\\ Process\end{tabular}} & \textbf{ACL} & \textbf{AAAI} & \textbf{IEEE} & \textbf{ACM} & \textbf{PubMed} \\ 
\midrule
\begin{tabular}[c]{@{}l@{}}Database\\ Search\end{tabular}            & 68           & 30            & 52            & 280          & 104             \\
\addlinespace
\begin{tabular}[c]{@{}l@{}}Title\\ Screening\end{tabular}              & 26           & 16            & 39            & 137          & 84              \\
\addlinespace
\begin{tabular}[c]{@{}l@{}}Abstract\\ Screening\end{tabular}           & 9            & 4             & 31            & 45           & 68              \\
\addlinespace
\begin{tabular}[c]{@{}l@{}}Full-Text\\ Screening\end{tabular}          & \textbf{9}            & \textbf{4}             & \textbf{20}            & \textbf{40}           & \textbf{63}              \\ 
\addlinespace
\begin{tabular}[c]{@{}l@{}}Model /\\ Experiment\end{tabular}          & \textbf{6}            & \textbf{3}             & \textbf{15}            & \textbf{35}           & \textbf{43}              \\
\bottomrule
\end{tabular}%
}
    \caption{Steps in the screening process and the number of papers retained in each database.} 
    \label{tbl:screening}
\end{table}

\subsection{Screening Process}

For subsequent steps in the screening process, we adhere to a set of defined inclusion criteria. Specifically, we include a paper if it met the following conditions for a focused and relevant review of the literature that aligns with the objectives of our study:
\begin{itemize}[noitemsep,topsep=0pt]
    \item Primarily focused on CAs irrespective of modality, such as text, speech, or embodied.
    \item Related to mental health and well-being. These could be related to depression, PTSD, or other conditions defined in the DSM-IV~\cite{bell1994dsm} or other emotion-related intervention targets such as stress. 
    \item Contribute towards directly improving mental health CAs. This could be proposing novel models or conducting user studies. 
\end{itemize}

The initial step in our screening process is title screening, in which we examine all titles, retaining those that are related to either CA or mental health. Our approach is deliberately inclusive during this phase to maximize the recall. As a result, out of 534 papers, we keep 302 for the next step.

Following this, we proceed with abstract screening. In this stage, we evaluate whether each paper meets our inclusion criteria. 
To enhance the accuracy and efficiency of our decision-making process, we extract the ten most frequent words from the full text of each paper to serve as keywords. These keywords provide an additional layer of verification, assisting our decision-making process. Following this step, we are left with a selection of 157 papers. 

The final step is full-text screening. When we verify if a paper meets the inclusion criteria, we extract key features (such as model techniques and evaluations) from the paper and summarize them in tables (see appendix). Simultaneously, we highlight and annotate the papers' PDF files to provide evidence supporting our claims about each feature similar to the methodology used in \citet{howcroft-etal-2020-twenty}. This process is independently conducted by two co-authors on a subset of 25 papers, and the annotations agree with each other. Furthermore, the two co-authors also agree upon the definition of features, following which all the remaining papers receive one annotation.\footnote{Annotated PDF files with evidence of each feature are available in our GitHub.}

The final corpus contains 136 papers: 9 from ACL, 4 from AAAI, 20 from IEEE, 40 from ACM, and 63 from PubMed. We categorize these papers into four distinct groups: 102 model/experiment papers, 20 survey papers, and the remaining 14 papers are classified as `other'. Model papers are articles whose primary focus is on the construction and explanation of a theoretical model, while experimental papers are research studies that conduct specific experiments on the models to answer pertinent research questions. We combine experiment and model papers together because experimental papers often involve testing on models, while model papers frequently incorporate evaluations through experiments. The `other' papers include dataset papers, summary papers describing the proceedings of a workshop, perspectives/viewpoint papers, and design science research papers. In this paper, we focus on analyzing the experiment/model and survey papers, which have a more uniform set of features.

\subsection{Feature Extraction}
We extract a set of 24 features to have a detailed and complete overview of the recent trend. They include general features (\textit{``paper type'', ``language'', ``mental health category'', ``background'', ``target group'', ``target demographic''}), techniques (\textit{``chatbot name'', ``chatbot type'', ``model technique'', ``off the shelf'', ``outsourced model name'', ``training data''}), appearance (\textit{``interface'', ``embodiment'', ``platform'', ``public access''}), and experiment (\textit{``study design'', ``recruitment'', ``sample size'', ``duration'', ``automatic evaluation'', ``human evaluation'', ``statistical test'', ``ethics''}). Due to the limited space, we present a 
subset of the features in the main paper. Description of other features can be found in Appendix.\footnote{Full feature table is available in the supplemental material.}

\begin{table}[t]
\centering
\resizebox{0.7\linewidth}{!}{%
\begin{tabular}{lccc}
\hline
\textbf{Language}  & \textbf{CS} & \textbf{Med} & \textbf{All} \\ \hline
English  & 47 & 30  & 77     \\
Chinese  & 1  & 5   & 6      \\
Korean   & 4  & 1   & 5      \\
German   & 1  & 1   & 2      \\
Italian  & 1  & 1   & 2      \\
Portuguese & 0  & 2 & 2      \\
Other   & 5  & 3   & 8      \\ \hline
\end{tabular}%
}
    \caption{Distribution of predominant language of the data and/or participants recruited in mental health CA papers. Other languages include Bangla, Danish, Dutch, Japanese, Kazakh, Norwegian, Spanish, and Swedish.}
    \label{tbl:language}
\end{table}

\section{Results}
Under the category of model and experiment papers, there are 6 papers from ACL, 3 from AAAI, 15 from IEEE, 35 from ACM, and 43 from PubMed. In this section, we briefly summarize the observations from the different features we extracted. 

\subsection{Language}
We identify if there is a predominant language associated with either the data used for the models or if there is a certain language proficiency that was a part of the inclusion criteria for participants. 
Our findings, summarized in Table \ref{tbl:language}, reveal that English dominates these studies with over 71\% of the papers utilizing data and/or participants proficient in English. Despite a few (17\%) papers emerging from East Asia and Europe, we notice that studies in low-resource languages are relatively rare. 

\begin{table}[]
\centering
\resizebox{\linewidth}{!}{%
\begin{tabular}{lccc}
\hline
\textbf{Mental Health Category} & \textbf{CS} & \textbf{Med} & \textbf{All} \\ \hline
Not Specified         & 32 & 21                & 53                        \\
Depression            & 9 & 10          & 19                        \\
Anxiety               & 8 & 8          & 16                        \\
Stress                & 0 & 4          & 4                         \\
Sexual Abuse          & 3 & 0               & 3                         \\
Social Isolation      & 3 & 0                 & 3                         \\
Other                 & 14 & 11         & 25                         \\ \hline
\end{tabular}%
}
    \caption{Distribution of mental health category in mental health CA papers. A paper could have multiple focused targets. Other categories include affective disorder, COVID-19, eating disorders, PTSD, substance use disorder, etc.}
    \label{tbl:mhcat}
\end{table}

\subsection{Mental Health Category}
Most of the papers (~43\%) we reviewed do not deal with a specific mental health condition but work towards general mental health well-being~\cite{saha-etal-2022-shoulder}. The methods proposed in such papers are applicable to the symptoms associated with a broad range of mental health issues (e.g. emotional dysregulation). Some papers, on the other hand, are more tailored to address the characteristics of targeted mental health conditions. As shown in Table \ref{tbl:mhcat}, depression and anxiety are two major mental health categories being dealt with, reflecting the prevalence of these conditions~\cite{eagle2022don}. Other categories include stress management \cite{park2019designing,gabrielli2021engagement}; sexual abuse, to help survivors of sexual abuse \cite{maeng2022designing, park2021designing}, and social isolation, mainly targeted toward older adults \cite{sidner2018creating, razavi2022discourse}. Less-studied categories include affective disorders~\cite{maharjan2022experiences, maharjan2022difference}, COVID-19-related mental health issues \cite{kim2022designing,ludin2022chatbot}, eating disorders \cite{beilharz2021development}, and PTSD \cite{han2021ptsdialogue}. 

\begin{table}[]
\centering
\resizebox{\linewidth}{!}{%
\begin{tabular}{lccc}
\hline
\textbf{Target Demographic} & \textbf{CS} & \textbf{Med} & \textbf{All} \\ \hline
General            & 43 & 26         & 69                        \\
Young People       & 4 & 6         & 10                        \\
Students           & 5 & 3         & 8                         \\
Women              & 3 & 4         & 7                         \\
Older adults       & 4 & 1         & 5                         \\
Other              & 1 & 4       & 5                         \\ \hline
\end{tabular}%
}
    \caption{Distribution of demographics focused by mental health CA papers. A paper could have multiple focused target demographic groups. Other includes black American, the military community, and employee.}
    \label{tbl:demographic}
\end{table}

\subsection{Target Demographic}
Most of the papers ($>$65\%) do not specify the target demographic of users for their CAs. The target demographic distribution is shown in Table \ref{tbl:demographic}. 
An advantage of the models proposed in these papers is that they could potentially offer support to a broad group of users irrespective of the underlying mental health condition. 
Papers without a target demographic and a target mental health category focus on proposing methods such as using generative language models for psychotherapy \cite{das2022conversational}, or to address specific modules of the CAs such as leveraging reinforcement learning for response generation \cite{saha2022shoulder}. On the other hand, ~31\% papers focus on one specific user group such as young individuals, students, women, older adults, etc, to give advanced assistance. Young individuals, including adolescents and teenagers, received the maximum attention~\cite{rahman2021adolescentbot}. Several papers also focus on the mental health care of women, for instance in prenatal and postpartum women~\cite{green2019expanding, chung2021chatbot} and sexual abuse survivors \cite{maeng2022designing, park2021designing}. Papers targeting older adults are mainly designed for companionship and supporting isolated elders \cite{sidner2018creating, razavi2022discourse}.

\begin{table}[!t]
\centering
\resizebox{0.9\linewidth}{!}{%
\begin{tabular}{lccc}
\hline
\textbf{Model Technique} & \textbf{CS} & \textbf{Med} & \textbf{All} \\ \hline
Retrieval-Based     & 27 & 22     & 49                        \\
Rule-Based          & 23 & 19     & 42                        \\
Generative          & 10 & 0     & 10                        \\
Not Specified       & 3 & 3     & 6                         \\ \hline
\end{tabular}%
}
\caption{Distribution of model techniques used in mental health CA papers. A paper could use multiple modeling techniques. The Not Specified group includes papers without a model but employing surveys to ask people's opinions and suggestions towards mental health CA.}
    \label{tbl:tech}
\end{table}

\subsection{Model Technique}
Development of Large Language Models such as GPT-series \cite{radford2019language, brown2020language} greatly enhanced the performance of generative models, which in turn made a significant impact on the development of CAs~\cite{das-etal-2022-conversational, nie2022conversational}. However, as shown in Table \ref{tbl:tech}, LLMs are yet to be utilized in the development of mental health CAs (as of the papers reviewed in this study), especially in medicine. No paper from PubMed in our final list dealt with generative models, with the primary focus being rule-based and retrieval-based CAs. 

Rule-based models operate on predefined rules and patterns 
such as if-then statements or decision trees to match user inputs with predefined responses. The execution of Rule-based CAs can be straightforward and inexpensive, but developing and maintaining a comprehensive set of rules can be challenging. Retrieval-based models rely on a predefined database of responses to generate replies. They use techniques like keyword matching \cite{daley2020preliminary}, similarity measures \cite{collins2022covid}, or information retrieval \cite{morris2018towards} to select the most appropriate response from the database based on the user's input. Generative model-based CAs are mostly developed using deep learning techniques such as recurrent neural networks (RNNs) or transformers, which learn from large amounts of text data and generate responses based on the learned patterns and structures. While they can often generate more diverse and contextually relevant responses compared to rule-based or retrieval-based models, they could suffer from hallucination and inaccuracies~\cite{azaria2023internal}.

\begin{table}[t]
\centering
\resizebox{0.9\linewidth}{!}{%
\begin{tabular}{lccc}
\hline
\textbf{Outsourced Model} & \textbf{CS} & \textbf{Med} & \textbf{All} \\ \hline
Google Dialogflow       & 11 & 2  & 13                        \\
Rasa                    & 5 & 5    & 10                        \\
Alexa                   & 4 & 0    & 4                         \\
DialoGPT                & 3 & 0    & 3                         \\
GPT                     & 3 & 0    & 3                         \\
X2AI                    & 0 & 3    & 3                         \\
Other                   & 17 & 6   & 23                        \\ \hline
\end{tabular}%
}
\caption[Caption for LOF]{Distribution of outsourced models used for building models used in mental health CA papers. Other includes Manychat\protect\footnotemark, Woebot \cite{fitzpatrick2017delivering} and Eliza \cite{weizenbaum1966eliza}.}
    \label{tbl:outsourced}
\end{table}

\subsection{Outsourced Models}
Building a CA model from scratch could be challenging for several reasons such as a lack of sufficient data, compute resources, or generalizability. 
Publicly available models and architectures have made building CAs accessible. Google Dialogflow \cite{Dialogflow} and Rasa \cite{bocklisch2017rasa} are the two most used outsourced platforms and frameworks. Alexa, DialoGPT \cite{zhang2019dialogpt}, GPT (2 and 3) \cite{radford2019language,brown2020language} and X2AI (now called Cass) \cite{X2AI} are also frequently used for building CA models. A summary can be found in Table \ref{tbl:outsourced}. 

\footnotetext{\url{https://manychat.com}}

Google Dialogflow is a conversational AI platform developed by Google that enables developers to build and deploy chatbots and virtual assistants across various platforms. Rasa is an open-source conversational AI framework that empowers developers to create and deploy contextual chatbots and virtual assistants with advanced natural language understanding capabilities. Alexa is a voice-controlled virtual assistant developed by Amazon. It enables users to interact with a wide range of devices and services using voice commands, offering capabilities such as playing music, answering questions, and providing personalized recommendations. DialoGPT is a large, pre-trained neural conversational response generation model that is trained on the GPT2 model with 147M conversation-like exchanges from Reddit. X2AI is the leading mental health AI assistant that supports over 30M individuals with easy access. 

\subsection{Evaluation}
\paragraph{Automatic:} Mental health CAs are evaluated with various methods and metrics. Multiple factors, including user activity (total sessions, total time, days used, total word count), user utterance (sentiment analysis, LIWC \cite{liwc2015}), CA response quality (BLEU \cite{papineni-etal-2002-bleu}, ROUGE-L \cite{lin-2004-rouge}, lexical diversity, perplexity), and performance of CA's sub-modules (classification f1 score, negative log-likelihood) are measured and tested. We find that papers published in the CS domain focus more on technical evaluation, while the papers published in medicine are more interested in user data.

\paragraph{Human outcomes:} Human evaluation using survey assessment is the most prevalent method to gauge mental health CAs' performance. 
Some survey instruments measure the pre- and post-study status of participants and evaluate the impact of the CA by comparing mental health  (e.g. PHQ-9~\cite{kroenke2001phq}, GAD-7~\cite{spitzer2006brief}, BFI-10~\cite{rammstedt2013short}) and mood scores (e.g. WHO-5~\cite{topp20155}), or collecting user feedback on CA models (usability, difficulty, appropriateness), or asking a group of individuals to annotate user logs or utterances to collect passive feedbacks (self-disclosure level, competence, motivational). 

\subsection{Ethical Considerations}
Mental health CAs inevitably work with sensitive data, including demographics, Personal Identifiable Information (PII), and Personal Health Information (PHI). Thus, careful ethical consideration and a high standard of data privacy must be applied in the studies. Out of the 89 papers that include human evaluations, approximately 70\% (62 papers) indicate that they either have been granted approval by Institutional Review Boards (IRB) or ethics review committees or specified that ethical approval is not a requirement based on local policy. On the other hand, there are 24 papers that do not mention seeking ethical approval or consequent considerations in the paper. Out of these 24 papers that lack a statement on ethical concerns, 21 papers are published in the field of CS. 

\section{Discussion}

\subsection{Disparity in Research Focus}
Mental health Conversational Agents require expert knowledge from different domains. However, the papers we reviewed, treat this task quite differently, evidenced by the base rates of the number of papers matching our inclusion criteria. 
For instance, there are over 28,000 articles published in the ACL Anthology with the keywords ``chatbot'' or ``conversational agent'', which reveals the popularity of this topic in the NLP domain. However, there are only 9 papers related to both mental health and CA, which shows that the focus of NLP researchers is primarily concentrated on the technical development of CA models, less on its applications, including mental health. AAAI shares a similar trend as ACL. However, there are a lot of related papers to mental health CAs in IEEE and ACM, which show great interest from the engineering and HCI community. PubMed represents the latest trend of research in the medical domain, and it has the largest number of publications that fit our inclusion criteria. While CS papers mostly do not have a specific focus on the mental health category for which CAs are being built, papers published in the medical domain often tackle specific mental health categories. 

\subsection{Technology Gap}
CS and medical domains are also different in the technical aspects of the CA model. In the CS domain (ACL, AAAI, IEEE, ACM), 41 (of 73 papers) developed CA models, while 14 (out of 63) from the medical domain (PubMed) developed models. Among these papers, 8 from the CS domain are based on generative methods, but no paper in PubMed uses this technology. The NLP community is actively exploring the role of generative LLMs (e.g. GPT-4) in designing CAs including mental healthcare-related CAs \cite{das2022conversational, saha2022shoulder, yan2021grounded}. With the advent of more sophisticated LLMs, \textit{fluency}, \textit{repetitions} and, \textit{ungrammatical formations} are no longer concerns for dialogue generation. However, stochastic text generation coupled with black box architecture prevents wider adoption of these models in the health sector \cite{vaidyam2019chatbots}. Unlike task-oriented dialogues, mental health domain CAs predominantly involve unconstrained conversation style for \textit{talk-therapy} that can benefit from the advancements in LLMs \cite{abd2021perceptions}. 

PubMed papers rather focus on retrieval-based and rule-based methods, which are, arguably, previous-generation CA models as far as the technical complexity is concerned. This could be due to a variety of factors such as explainability, accuracy, and reliability which are crucial when dealing with patients.

\subsection{Response Quality vs Health Outcome}

The difference in evaluation also reveals the varying focus across CS and medicine domains. From the CS domains, 30 (of 59 papers) applied automatic evaluation, which checks both model's performance (e.g. BLEU, ROUGE-L, perplexity) and participant's CA usage (total sessions, word count, interaction time). In contrast, only 13 out of 43 papers from PubMed used automatic evaluation, and none of them investigated the models' performance. 

The difference is also spotted in human evaluation. 40 (of 43 papers) from PubMed consist of human outcome evaluation, and they cover a wide range of questionnaires to determine participants' status (e.g. PHQ-9, GAD-7, WHO-5). The focus is on users' psychological well-being and evaluating the chatbot's suitability in the clinical setup \cite{martinengo2022conversational}. Although these papers do not test the CA model's performance through automatic evaluation, they asked for participants' ratings to oversee their model's quality (e.g. helpfulness, System Usability Scale \cite{brooke1996sus}, WAI-SR \cite{munder2010working}).

All 6 ACL papers that satisfied our search criteria, solely focus on dialogue quality (e.g. \textit{fluency}, \textit{friendliness} etc.) with no discussion on CA's effect on users' well-being through clinical measures such as PHQ-9. CAs that aim to be the first point of contact for users seeking mental health support, should have clinically validated mechanisms to monitor the well-being of their users \cite{pacheco2021smart,wilson2022development}. Moreover, the mental health CAs we review are designed without any underlying theory for psychotherapy or behavior change that puts the utility of CAs providing \textit{emotional support} to those suffering from mental health challenges in doubt. 

\subsection{Transparency}
None of the ACL papers that we reviewed released their model or API. Additionally, a \textit{baseline} or comparison with the existing state-of-the-art model is often missing in the papers. There is no standardized outcome reporting procedure in both medicine and CS domains \cite{vaidyam2019chatbots}. For instance, \citet{valizadeh2022ai} raised concerns about the replicability of evaluation results and transparency for healthcare CAs. We acknowledge the restrictions posed to making the models public due to the sensitive nature of the data. However, providing APIs could be a possible alternative to enable comparison for future studies. To gauge the true advantage of mental health CAs in a clinical setup, randomized control trials are an important consideration that is not observed in NLP papers. Further, standardized benchmark datasets for evaluating mental health CAs could be useful in increasing transparency. 

\subsection{Language and Cultural Heterogeneity}
Over 75\% of the research papers in our review cater to English-speaking participants struggling with depression and anxiety. Chinese and Korean are the two languages with the highest number of research papers following English, even though Chinese is the most populous language in the world. Future works could consider tapping into a diverse set of languages that also have a lot of data available - for instance, Hindi, Arabic, French, Russian, and Japanese, which are among the top 10 most spoken languages in the world. The growing prowess of multilingual LLMs could be an incredible opportunity to transfer universally applicable development in mental health CAs to low-resource languages while being mindful of the racial and cultural heterogeneity which several multilingual models might miss due to being trained on largely English data~\cite{bang2023multitask}. 

\section{Conclusion}
In this paper, we used the PRISMA framework to systematically review the recent studies about mental health CA across both CS and medical domains. From the well-represented databases in both domains, we begin with 865 papers based on a keyword search to identify mental health-related conversational agent papers and use title, abstract, and full-text screening to retain 136 papers that fit our inclusion criteria. Furthermore, we extract a wide range of features from model and experiment papers, summarizing attributes in the fields of general features, techniques, appearance, and experiment. Based on this information, we find that there is a gap between CS and medicine in mental health CA studies. They vary in research focus, technology, and evaluation purposes. We also identify common issues that lie between domains, including transparency and language/cultural heterogeneity. 

\section*{Potential Recommendations}
We systematically study the difference between domains and show that learning from each other is highly beneficial. Since interdisciplinary works consist of a small portion of our final list (20 over 102 based on author affiliations on papers; 7 from ACM, 2 from IEEE, and 11 from PubMed), we suggest more collaborations to help bridge the gap between the two communities. For instance, NLP (and broadly CS) papers on mental health CAs would benefit from adding pre-post analysis on human feedback and considering ethical challenges by requesting a review of an ethics committee. Further, studies in medicine could benefit by tapping into the latest developments in generative methods in addition to the commonly used rule-based methods. In terms of evaluation, both the quality of response by the CAs (in terms of automatic metrics such as BLEU, ROUGE-L, perplexity, and measures of dialogue quality) as well as the effect of CA on users' mental states (in terms of mental health-specific survey inventories) could be used to assess the performance of mental health CAs. Moreover, increasing the language coverage to include non-English data/participants and adding cultural heterogeneity while providing APIs to compare against current mental health CAs would help in addressing the challenge of mental health care support with a cross-disciplinary effort.





\section*{Limitations}
This survey paper has several limitations. Our search criteria are between January 2017 to December 2022, which likely did not reflect the development of advanced CA and large language models like ChatGPT and GPT4 \cite{sanderson2023gpt}. We couldn't include more recent publications to meet the EMNLP submission date. Nonetheless, we have included relevant comments across the different sections on the applicability of more sophisticated models. 

Further, search engines (e.g. Google Scholar) are not deterministic. Our search keywords, filters, and chosen databases do not guarantee the exact same search results. However, we have tested multiple times on database searching and they returned consistent results. We have downloaded PDFs of all the papers and have saved the annotated them to reflect the different steps used in this review paper. These annotations will be made public.

For some databases, the number of papers in the final list may be (surprisingly!) small to represent the general research trends in the respective domains. However, it also indicates the lack of focus on mental health CA from these domains, which also proposes further attention is required in the field.

\section*{Ethics Statement}
Mental Health CAs, despite their accessibility, potential ability, and anonymity, cannot replace human therapists in providing mental health care. There are a lot of ongoing discussions about the range of availability of mental health CAs, and many raise several challenges and suspicions about automated conversations. Rule-based and retrieval-based models can be controlled for content generation, but cannot answer out-of-domain questions. Generative models are still a developing field, their non-deterministic nature raises concerns about the safety and reliability of the content. Thus at the current stage, CA could play a great supporting complementary role in mental healthcare to identify individuals who potentially need more immediate care in an already burdened healthcare system.

Since the patient's personal information and medical status are extremely sensitive, we highly encourage researchers and developers to pay extra attention to data security and ethics ~\citet{arias2022quantifying}. The development, validation, and deployment of mental health CAs should involve multiple diverse stakeholders to determine how, when, and which data is being used to train and infer participants' mental health. This effort requires a multidisciplinary effort to address the complex challenges of mental health care~\cite{chancellor2019taxonomy}.

\section*{Acknowledgements}

We would like to thank the reviewers for their fruitful discussion with us. This work was partly supported by grant NIMHD: R01MD018340 from the National Institutes of Health and Penn Global Research Engagement Fund. The funders had no role in the design and conduct of the study; collection, management, analysis, and interpretation of the data; preparation, review, or approval of the manuscript; and decision to submit the manuscript for publication.



\bibliography{anthology,custom}
\bibliographystyle{acl_natbib}

\appendix

\begin{table*}[b!]
\centering
\scriptsize
\begin{tabular}{llllllllll}
\hline
\multicolumn{2}{c}{\textbf{AAAI}} & \multicolumn{2}{c}{\textbf{ACL}} & \multicolumn{2}{c}{\textbf{ACM}} & \multicolumn{2}{c}{\textbf{IEEE}} & \multicolumn{2}{c}{\textbf{PubMed}} \\
\hline
Venue           & Count           & Venue            & Count         & Venue                & Count     & Venue                    & Count  & Venue                     & Count   \\ \hline
HCOMP           & 2               & EMNLP            & 1             & CHI                  & 9         & ICIRCA                   & 2      & JMIR Form Res             & 9       \\
AAAI            & 1               & SIGDIAL          & 1             & ACM-TiiS             & 4         & ACII                     & 2      & J Med Internet Res        & 7       \\
                &                 & BioNLP           & 1             & IVA                  & 4         & ICoICT                   & 1      & Front Digit Health        & 4       \\
                &                 & NAACL            & 1             & ACM-HCI              & 3         & UCET                     & 1      & JMIR Mhealth Uhealth      & 3       \\
                &                 & NLP4PI           & 1             & UbiComp-ISWC         & 2         & ICCCI                    & 1      & JMIR Res Protoc           & 3       \\
                &                 & LREC             & 1             & CUI                  & 2         & ICHCI                    & 1      & Digit Health              & 2       \\
                &                 &                  &               & PervasiveHealth      & 2         & ICACCS                   & 1      & JMIR Ment Health          & 2       \\
                &                 &                  &               & CHItaly              & 1         & ISCC                     & 1      & JMIR Hum Factors          & 2       \\
                &                 &                  &               & ACSW                 & 1         & IEEE Trans. Emerg.       & 1      & Internet Interv           & 2       \\
                &                 &                  &               & H3                   & 1         & SIEDS                    & 1      & Curr Psychol              & 1       \\
                &                 &                  &               & Asian CHI            & 1         & IEEE Pervasive Comput.   & 1      & Comput Math Methods Med   & 1       \\
                &                 &                  &               & DIS                  & 1         & ICCAS                    & 1      & Inf Process Manag         & 1       \\
                &                 &                  &               & CHIuXiD              & 1         & INCET                    & 1      & Front Psychol             & 1       \\
                &                 &                  &               & ACM-HEALTH           & 1         &                          &        & Trials                    & 1       \\
                &                 &                  &               & IASA                 & 1         &                          &        & Front Psychiatry          & 1       \\
                &                 &                  &               & ECCE                 & 1         &                          &        & Drug Alcohol Depend       & 1       \\
                &                 &                  &               &                      &           &                          &        & Sensors (Basel)           & 1       \\
                &                 &                  &               &                      &           &                          &        & JMIR Med Inform           & 1       \\ \hline
\end{tabular}
    \caption{Venues in each database that have at least one paper in our final list and the corresponding number of model/experiment papers. }
    \label{tab:subvenue}
\end{table*}

\section{Venues of Selected Papers}
In this paper, we searched all venues indexed under 5 databases to cover most of the venues that are interested in mental health conversational agents. In Table \ref{tab:subvenue}, we show the distribution of venues under each database for the papers that are selected for the final list. 

\section{Full Table Explanation}
\label{sec:appendix}

We show our final list of model/experiment papers in Table \ref{tab:final_general_appearance}, Table \ref{tab:final_techniques} and Table \ref{tab:final_experiment}. Due to the limited size of the paper, some columns (``background'') are removed and long values are truncated. The full table is available on our GitHub.

For an easier understanding of our full table, we briefly introduce each feature we extracted below.
\begin{itemize}[noitemsep,topsep=0pt]
    \item \textit{Paper}: The citation of the selected paper.
    \item \textit{Database}: The source of the paper.
    \item \textit{Paper Type}: The type of the paper. We here only show model or experiment papers.
    \item \textit{Language}: Target language used in this paper.
    \item \textit{Mental Health Category}: Target mental health category in this paper.
    \item \textit{Target Group}: Target group of this paper. Could be patients, caregivers, or clinicians.
    \item \textit{Target Demographic}: Target demographic of this paper. If it is not specified or can be used by anyone, we mark it as General.
    \item \textit{Chatbot Name}: The name of the chatbot model used in this paper.
    \item \textit{Chatbot Type}: Type of the mental health CA. Could be QA, open domain, or task-oriented.
    \item \textit{Model Technique}: Type of technique used to build the model. Could be rule-based, retrieval-based, or generative.
    \item \textit{Off the Shelf}: Information about the usage of off-the-shelf models in the system. We limit Off-the-shelf models to pre-trained models or applications. Could be yes (directly used), used as a part (off-the-shelf model is a part of the pipeline), or finetuned.
    \item \textit{Outsourced Model Name}: The name of the off-the-shelf model, if any.
    \item \textit{Training Data}: The name or source of the training data, if any.
    \item \textit{Interface}: Type of input the model takes. Could be text, voice, visual, or button.
    \item \textit{Embodiment}: Embodiment of the model. Could be physical or visual.
    \item \textit{Platform}: The platform the model run on. Could be Web, Mobile, PC, or other devices.
    \item \textit{Public Access}: If the availability of the model is disclosed in the paper. Could be fully open (parameter level) or API (able to use).
    \item \textit{Study Design}: Type of user study performed in the paper. Could be RCT (Randomized Controlled Trial), user study (ask participants to use and evaluate), or comparative analysis (divide users with different conditions and compare the results).
    \item \textit{Recruitment}: How participants are recruited.
    \item \textit{Sample Size}: Size of the participants.
    \item \textit{Duration}: Duration of the user study.
    \item \textit{Automatic Evaluation}: List of automatic evaluation metrics used in this paper.
    \item \textit{Human Evaluation}: List of parameters/metrics derived from Human Evaluation used in this paper.
    \item \textit{Statistical Test}: List of statistical tests used for measuring significance in this paper.
    \item \textit{Ethics}: Whether the paper mentioned ethical consideration. Could be IRB (Institutional Review Board), or yes (ethical consideration is mentioned in the paper).
\end{itemize}

\clearpage
\onecolumn
\begin{landscape}
\tiny
\begin{longtable}{lllllllllll}
\caption{All method/experiment papers in the final list of this survey. This table only shows general and appearance features.}
\label{tab:final_general_appearance}\\
\hline
\textbf{Paper}                     & \textbf{Database} & \textbf{Paper Type} & \textbf{Language} & \textbf{Mental Health Category}                  & \textbf{Target Group} & \textbf{Target Demographic}  & \textbf{Interface} & \textbf{Embodiment} & \textbf{Platform}       & \textbf{Public Access} \\ \hline
\endhead
\hline
\endfoot
\endlastfoot
\citet{abbas2020trainbot}          & AAAI              & Experiment          & English           & General                                          & Clinicians            & General                      & Text               & /                   & Web                     & /                      \\
\citet{sun2022comparing}           & AAAI              & Experiment          & English           & General                                          & Clinicians            & General                      & Text               & /                   & Web                     & /                      \\
\citet{garg2020modeling}           & AAAI              & Model               & English           & Depression                                       & Patients              & General                      & Text               & /                   & /                       & /                      \\
\citet{ishii2021erica}             & ACL               & Experiment          & English           & Isolation                                        & Patients              & General                      & Voice, Visual      & Physical            & Web                     & /                      \\
\citet{demasi2020multi}            & ACL               & Model               & English           & Suicide                                          & Clinicians            & General                      & Text               & /                   & /                       & /                      \\
\citet{das2022conversational}      & ACL               & Model               & English           & General                                          & Patients              & General                      & Text               & /                   & /                       & /                      \\
\hline
\end{longtable}
\end{landscape}

\begin{landscape}
\scriptsize
\begin{longtable}{lllllll}
\caption{All method/experiment papers in the final list of this survey. This table only shows technique features. Long values are truncated due to limited space.}
\label{tab:final_techniques}\\
\hline
\textbf{Paper}                     & \textbf{Chatbot Name}             & \textbf{Chatbot Type} & \textbf{Model Technique}    & \textbf{Off the Shelf}    & \textbf{Outsourced Model Name}      & \textbf{Training Data}            \\ \hline
\endhead
\hline
\endfoot
\endlastfoot
\citet{abbas2020trainbot}          & Trainbot                          & Task Oriented         & Rule-Based                  & /                         & /                                   & /                                 \\
\citet{sun2022comparing}           & MemberBot                         & QA                    & Retrieval-Based             & Used As a Part            & Rasa                                & (New) 7cups Conversation Data     \\
\citet{garg2020modeling}           & Unnamed                           & Open Domain           & Retrieval-Based             & /                         & /                                   & Depression Therapy Sessions, L... \\
\citet{ishii2021erica}             & ERICA, Nora                       & Task Oriented         & Rule-Based                  & Used As a Part            & Nora                                & /                                 \\
\citet{demasi2020multi}            & Crisisbot                         & Task Oriented         & Generative, Retrieval-Based & /                         & /                                   & Realistic Hotline Training Con... \\
\citet{das2022conversational}      & GPT2, DIALOGPT                    & Open Domain           & Generative                  & Finetuned                 & GPT-2, DIALOGPT                     & (New) Reddit, Transcripts Of A... \\
\citet{saha2022shoulder}           & MIC Model                         & Open Domain           & Generative                  & Finetuned, Used As a Part & DialoGPT                            & (New) MotiVAte                    \\
\citet{yan2021grounded}            & SocialBot, Chatbot                & Open Domain           & Retrieval-Based, Generative & Finetuned, Used As a Part & GPT                                 & (New) Medline Data, MedDialog,... \\
\hline
\end{longtable}
\end{landscape}



\begin{landscape}
\tiny
\begin{longtable}{lllllllll}
\caption{All method/experiment papers in the final list of this survey. This table only shows experiment features. Long values are truncated due to limited space.}
\label{tab:final_experiment}\\
\hline
\textbf{Paper}                     & \textbf{Study Design} & \textbf{Recruitment}              & \multicolumn{1}{l}{\textbf{Sample Size}} & \textbf{Duration} & \textbf{Automatic Evaluation}     & \textbf{Human Evaluation}         & \textbf{Ethics} & \textbf{Statistical Test}         \\ \hline
\endhead
\hline
\endfoot
\endlastfoot
\citet{abbas2020trainbot}          & Comparative Analysis  & Prolific.Ac                       & \multicolumn{1}{l}{100, 100}             & /                 & /                                 & Enjoyment, Pressure, Helping S... & Yes             & Independent Samples T-Test, Tw... \\
\citet{sun2022comparing}           & User Study            & MTurk Workers And Domain Exper... & \multicolumn{1}{l}{15, 11}               & /                 & Number Of Messages, Length Of ... & Difficulty, Enjoyment             & /               & Mann–Whitney U Test, Linear Re... \\
\citet{garg2020modeling}           & /                     & /                                 & \multicolumn{1}{l}{/}                    & /                 & Alignment                         & Appropriate, Diverse              & /               & /                                 \\
\citet{ishii2021erica}             & Comparative Analysis  & Recruited                         & 19                                       & /                 & /                                 & Overall Experience, Empathy, A... & /               & One-Sided t-Test                  \\
\citet{demasi2020multi}            & User Study            & Crowdworkers And Counslers        & \multicolumn{1}{l}{30, 5}                & /                 & Negative Log Likelihood, Entro... & Coherency, Consistency, Fluenc... & IRB             & /                                 \\
\citet{das2022conversational}      & User Study            & Psychiatrist, Psychologist        & \multicolumn{1}{l}{1,1}                  & /                 & Lexical Diversity, Average Cos... & Communication, Basic Psychothe... & /               & Cohen’s x, Krippendorff’s $\alpha$       \\
\hline
\end{longtable}
\end{landscape}

\end{document}